\documentclass[conference,letterpaper]{IEEEtran}
\IEEEoverridecommandlockouts
\usepackage[T1]{fontenc}
\usepackage{cite,amsmath,amssymb,graphicx,booktabs,array,tikz,xcolor}
\usetikzlibrary{arrows.meta,positioning,fit}
\usepackage[hidelinks]{hyperref}
\usepackage{url,flushend}
\usepackage{subcaption}
\definecolor{pendingcolor}{RGB}{135,53,22}

\newcommand{\rt}{\texttt{routine}}
\newcommand{\sd}{\texttt{see-doctor}}
\newcommand{\ur}{\texttt{urgent}}
\hypersetup{pdftitle={HerHealthGPT: An Evidence-Grounded Conversational AI System for Women's Reproductive Health},pdfsubject={ICCE 2027 draft}}
\begin{document}
\title{HerHealthGPT: An Evidence-Grounded Conversational AI System for Women's Reproductive Health%
\thanks{The source code and implementation details are publicly available at: \url{https://github.com/HerHealthGPT/herhealthgpt}.}
\thanks{This work has been submitted to the IEEE for possible publication. Copyright may be transferred without notice, after which this version may no longer be accessible.}
}
\author{\IEEEauthorblockN{Hassan Saeed Hassan Albattra, Mazen Mohammed Bahgat, Rahatara Ferdousi,\\
Hana Essam Sayed Ahmed Amrya, and Mariam Mousa}
\IEEEauthorblockA{School of Computing, Queen's University, Kingston, Ontario, Canada\\
\{hassan.albattra, mazen.bahgat, rahatara.ferdousi, hana.amrya, mariam.mousa\}@queensu.ca}}
\maketitle

\begin{abstract}
“Why is my period late?” Sometimes, the most useful response begins with another question. Yet conversational AI systems may respond to incomplete health queries without seeking essential context or grounding their guidance in verifiable evidence. This paper introduces HerHealthGPT, an evidence-grounded conversational AI system designed to address these challenges in women's reproductive health. HerHealthGPT integrates retrieval-augmented generation (RAG), context-aware clarification, and source validation within an interactive web interface. Rather than responding immediately to incomplete queries, the system maintains conversational context, asks targeted follow-up questions, checks for premature conclusions, and grounds its responses in retrieved evidence. We evaluate the AI components using 540 English messages derived from 90 reproductive-health cases and conduct a prototype questionnaire study with 30 adults. Evidence retrieval improves concern-category accuracy from 54.1\% to 59.3\% compared with fine-tuning alone. The clarification router achieves 85.6\% recall and 86.2\% specificity on constructed targets. In the user study, 83.3\% of participants preferred the retrieval-enabled configuration, while 25 of 30 considered its follow-up questions relevant. These findings suggest that integrating evidence retrieval with interactive clarification offers a promising approach to more transparent, context-aware reproductive-health information systems, while further clinical validation remains necessary.
\end{abstract}


\section{Introduction}
\label{sec:introduction}

A woman experiencing irregular periods may ask an AI
assistant for guidance without providing details about
symptom frequency, duration, or severity \cite{menst2025}. A conventional
chatbot may respond immediately, even when important
information is missing. A reliable consumer-health
assistant should instead recognize uncertainty, ask
relevant follow-up questions, and support its responses
with verifiable evidence. This is particularly important
in women's reproductive health, where concerns may be
expressed using clinical terminology, everyday language,
or indirect and emotionally sensitive descriptions
\cite{zeng2006}.

Large language models (LLMs) have enabled new approaches
to consumer-health assistance. UbiMyTherapist combines
contextual information and AI inference for mental-health
support \cite{alghoul2026}. Related studies investigate
generative AI for mental-health experiences \cite{yu2024},
accessible conversational interfaces \cite{alofi2026},
and domain-specialized menstrual-health models
\cite{menst2025}, while RHealthTwin explores
structured consumer-health interpretation with retrieval
grounding \cite{ferdousi2025}. Although these systems demonstrate
the potential of LLMs, providing relevant information
is only part of the challenge. An assistant must also
determine when to ask questions, provide guidance,
or defer a conclusion.

Retrieval-augmented generation (RAG) can improve evidence
grounding \cite{lewis2020}, while parameter-efficient
fine-tuning supports domain adaptation \cite{dettmers2023}.
However, neither approach independently ensures that
sufficient information has been collected before
responding. Research on medical retrieval
\cite{xiong2024} and clarification strategies
\cite{testoni2024,zhang2025,li2024} highlights the
importance of evidence selection, follow-up questioning,
and uncertainty management. Integrating these capabilities
into a practical consumer-facing system remains an
important challenge.

To address this gap, we present \textit{HerHealthGPT},
an evidence-grounded conversational AI system for
women's reproductive-health information. The system
combines domain-adapted language models, evidence
retrieval, structured conversational decisions, and
citation validation. Its web-based interface maintains
conversation history and supports targeted follow-up
questions before presenting source-linked guidance.
\begin{figure*}[!t]
    \centering
    \includegraphics[width=0.7\textwidth]
    {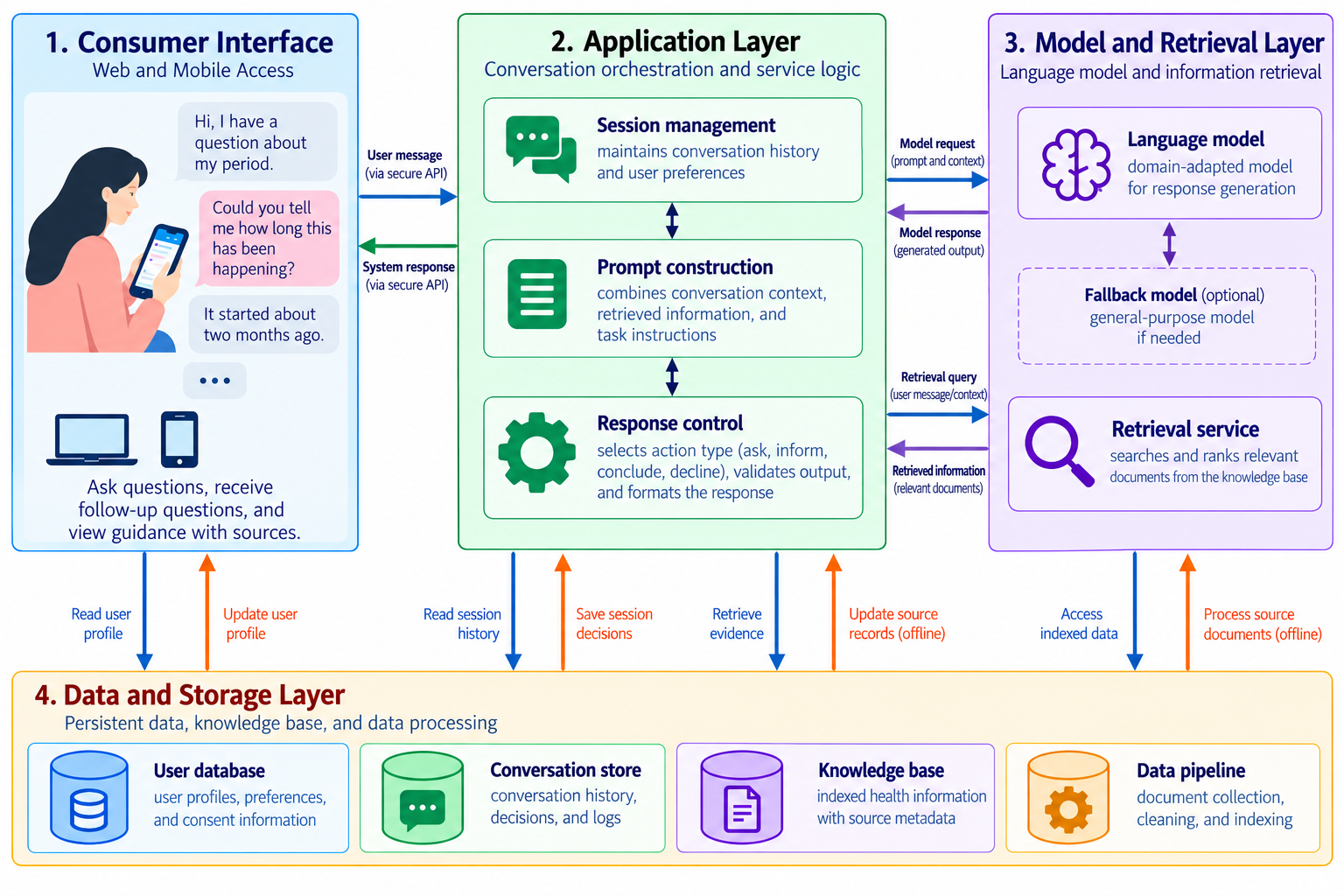}
    \caption{Four-layer architecture of HerHealthGPT,
    integrating the consumer interface, application services,
    model inference and retrieval, and persistent storage.}
    \label{fig:architecture}
\end{figure*}

The main contributions are:
\begin{itemize}
    \item \textbf{An evidence-grounded conversational
    system:} We develop HerHealthGPT, integrating
    domain adaptation, retrieval, session continuity,
    and source-traceable responses within a modular
    consumer-health architecture.

    \item \textbf{A structured conversational workflow:}
    We introduce targeted follow-up questioning,
    premature-conclusion checks, structured response
    decisions, and citation validation to support
    context-aware interactions.
\end{itemize}

We evaluate the approach through controlled component
experiments and a consumer questionnaire. On 540 English
messages \cite{herhealth2026}, retrieval improves concern-category accuracy
from 54.1\% to 59.3\% over fine-tuning alone. An
exploratory clarification router achieves 85.6\% recall
on constructed clarification targets. In a study with
30 adults, 83.3\% preferred the retrieval-enabled
configuration. These results support the value of
evidence grounding while identifying limitations in
urgency assessment and conversational reliability.

The remainder of this paper is organized as follows.
Section~\ref{sec:system} presents the HerHealthGPT architecture and
conversational workflow. Section~\ref{sec:evaluation} describes the
evaluation methodology. Section~\ref{sec:results} presents and discusses the experimental and consumer-study results.
Finally, Section~\ref{sec:conclusion} concludes the paper and outlines
future research directions.
\begin{figure*}[!t]
    \centering
    \includegraphics[width=0.92\textwidth]
    {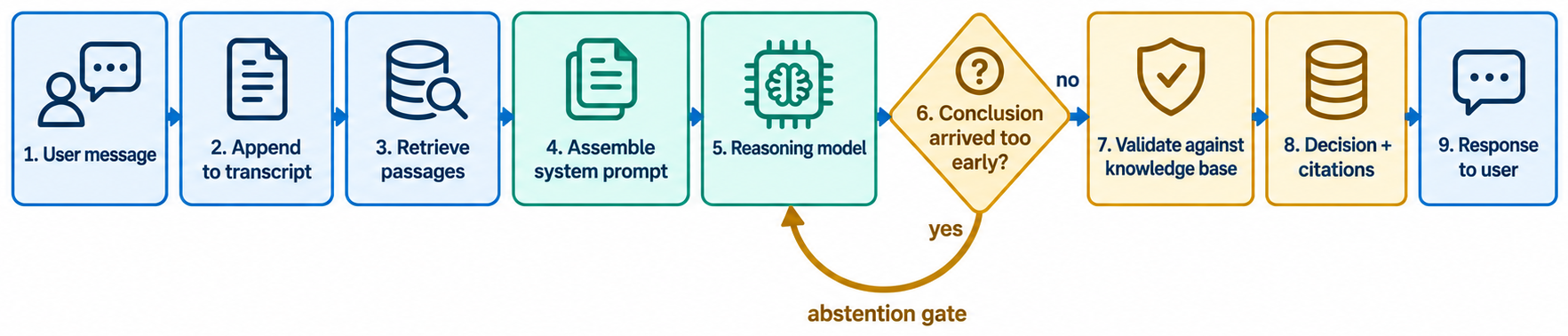}
    \caption{HerHealthGPT conversational workflow.
    Each turn combines conversation history and
    retrieved evidence to generate a structured
    decision. A premature-conclusion check may
    trigger one revision before validation,
    storage, and response delivery.}
    \label{fig:loop}
\end{figure*}

\section{System Architecture}
\label{sec:system}

We propose HerHealthGPT as an evidence-grounded conversational AI
system for women's reproductive-health information.
It integrates follow-up questioning, evidence retrieval,
and structured response generation through four layers:
consumer interface, application services, inference and
retrieval, and persistent storage (Fig.~\ref{fig:architecture}). We describe the architecture as follows.

\subsection{Consumer Interface}

The  web interface supports desktop and mobile
access, with English and Arabic right-to-left layouts.
Users receive information about the system's purpose
and limitations during onboarding. Account management
and persistent sessions allow conversations to be resumed.

The interface presents one follow-up question at a time
with supporting sources. Conclusions provide source-linked
guidance and possible health conditions as informational
patterns rather than diagnoses. Users may also request
general information or submit out-of-scope questions
without ending a session. Figure~\ref{fig:ui} illustrates
the main interactions.

\subsection{Application and Evidence Services}

A FastAPI backend manages sessions, prompt construction,
model requests, and response validation. PostgreSQL stores
conversation history, structured decisions, and citations.
Response generation uses Qwen3.5-9B \cite{qwen2026}.
The base model is accessed through a hosted API, while
the QLoRA-adapted model runs on a self-hosted vLLM server
\cite{kwon2023}. The application falls back to the hosted
model when the adapted model is unavailable.

The evidence store contains 39 askable features,
19 conditions, 146 evidence records, and 17 warning-sign
rules (seven emergency and ten same-day urgent),
developed from five public sources. Each evidence
record links a feature and condition to a traceable
source statement and evidence-strength label.
An absent relationship does not imply evidence
against a condition.

Retrieval uses BGE-M3 embeddings \cite{chen2024}
to identify relevant sentence-level passages.
It returns up to eight candidates, reserves three
positions for evidence statements, and excludes
matches below a cosine similarity of 0.3.
NHS content is excluded from the product retrieval
index for licensing reasons. This evidence store
is separate from the 49-chunk experimental corpus
used in the component evaluation.

\subsection{Evidence-Grounded Conversational Control}

HerHealthGPT generates structured decisions using
the current message, conversation history, and
retrieved evidence. Four response modes are supported:
\texttt{ask} for follow-up questions,
\texttt{conclude} for source-linked guidance,
\texttt{inform} for general information, and
\texttt{decline} for out-of-scope requests.
The latter two modes do not terminate the session.

Conclusions use four action codes:
\texttt{EMERGENCY}, \texttt{URGENT\_SAME\_DAY},
\texttt{ROUTINE\_APPOINTMENT}, and
\texttt{SELF\_CARE}. These product actions
are distinct from the three risk labels used
in the component evaluation.

As illustrated in Fig.~\ref{fig:loop}, each turn
combines the full conversation history with retrieved
evidence, controlled feature definitions, and
warning-sign information to generate a decision.
The application checks for premature conclusions
when a more severe warning-sign rule is one answer
away, or when fewer than three answers have been
collected and an informative question remains
without a triggered warning sign. An active warning
sign is not rejected solely because fewer than
three answers are available.

When a conclusion is considered premature, the
application requests one revised decision identifying
the missing information. The revision proceeds directly
to validation without further retries. The application
then checks response structure, controlled identifiers,
and citation references before storing and displaying
the decision. Model identity and failed attempts are
also recorded for traceability.

These controls support consistent response formatting
and evidence traceability, but do not guarantee
clinical correctness. In particular, warning-sign
definitions guide model-selected decisions rather
than enforcing deterministic escalation.

\section{Prototype and Evaluation}
\label{sec:evaluation}

\begin{figure*}[t]
\centering
\begin{minipage}[b]{0.13\textwidth}\centering
\includegraphics[height=46mm]{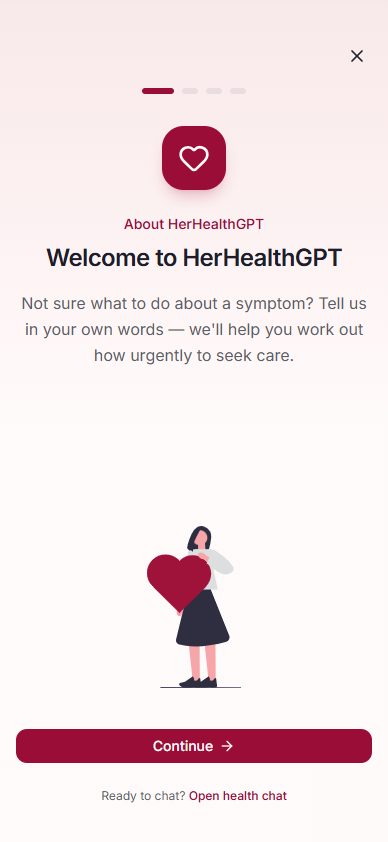}\\[-0.5mm]
{\footnotesize (a) Onboarding}
\end{minipage}\hfill
\begin{minipage}[b]{0.28\textwidth}\centering
\includegraphics[height=46mm]{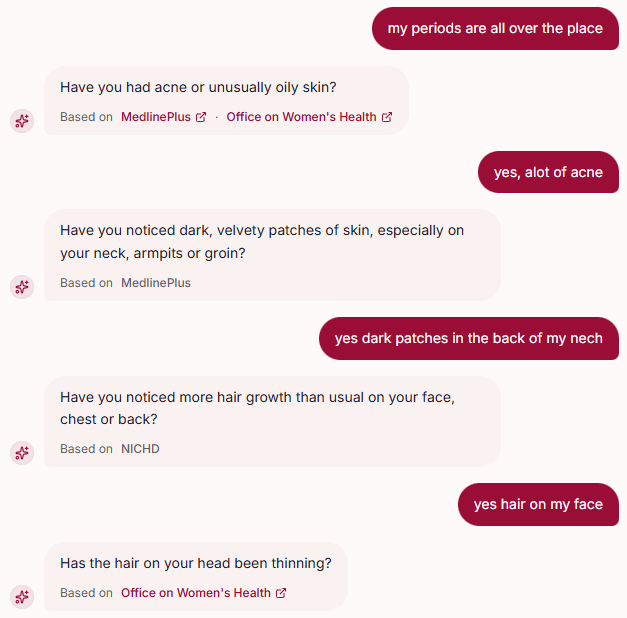}\\[-0.5mm]
{\footnotesize (b) Source-linked follow-up questions}
\end{minipage}\hfill
\begin{minipage}[b]{0.27\textwidth}\centering
\includegraphics[height=46mm]{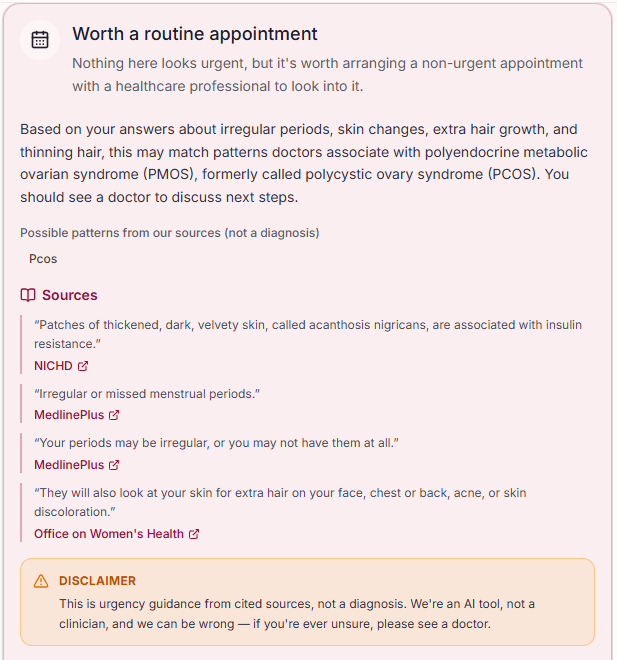}\\[-0.5mm]
{\footnotesize (c) Conclusion with quotations}
\end{minipage}\hfill
\begin{minipage}[b]{0.26\textwidth}\centering
\includegraphics[height=46mm]{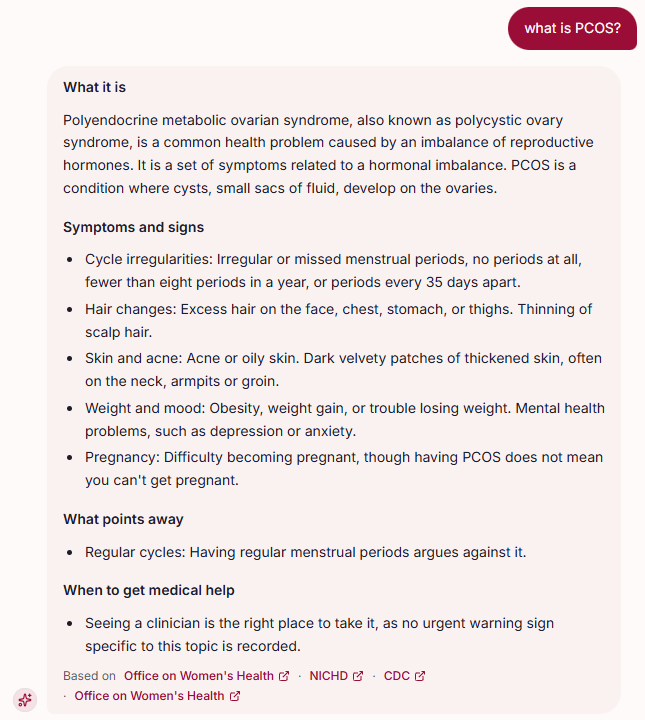}\\[-0.5mm]
{\footnotesize (d) Informational answer}
\end{minipage}
\caption[Prototype screens.]{Prototype screens.\protect\footnotemark[1]{} (a) Onboarding on a phone-sized screen; (b) one follow-up question per turn, each with its sources; (c) a routine-appointment conclusion with the quoted source sentences and disclaimer; (d) an \texttt{inform} answer to a general question, with citations and no urgency level. The examples show interface behavior, not clinical endorsements.}
\label{fig:ui}
\end{figure*}

We evaluate HerHealthGPT prototype (Fig.~\ref{fig:ui}) through controlled component
experiments and a consumer questionnaire. The former
examines model adaptation, retrieval, and clarification,
while the latter assesses user perceptions of the
implemented prototype. The experiment setup is described as followings.

\subsection{Component Evaluation}

\textbf{Benchmark and configurations:}
We use the English HerHealthEval benchmark
\cite{herhealth2026}, comprising 90 reproductive-health
cases expressed in six styles: canonical, clinical,
layperson, indirect, emotional, and under-specified
(540 messages). The benchmark includes 90 constructed
clarification-positive messages. Heuristic risk labels
identify 348 routine (\rt), 174 same-day (\sd),
and 18 urgent (\ur) messages.

We compare six Qwen3.5-9B configurations: zero-shot
(ZS), base with RAG, fine-tuned (FT), FT with
instructions, FT with shuffled evidence, and FT
with relevant evidence (FT+RAG). All adapted
configurations share a fixed QLoRA checkpoint,
trained on 10,538 multilingual records with
1,989 validation records. Training uses two epochs,
learning rate $2\times10^{-4}$, LoRA rank 16,
gradient accumulation 16, and seed 3407.
Evaluation is restricted to English.

\textbf{Retrieval and inference:}
The experimental corpus contains 49 NHS and NICHD
sections covering PCOS, heavy menstrual bleeding,
infertility, and endometriosis
\cite{nhs_pcos,nhs_heavy,nhs_infertility,
nichd_endo,nichd_infertility}.
BGE-M3 embeddings and FAISS normalized inner-product
search retrieve the top three passages
\cite{chen2024,douze2024}. The shuffled control
uses nonidentical passage bundles from other cases,
without guaranteeing semantic irrelevance.
All configurations use greedy decoding, 4-bit
loading, disabled thinking mode, and a 512-token
generation limit. Retrieval is precomputed,
so online latency is not evaluated.

\textbf{Metrics and statistical analysis:}
We measure concern-category accuracy, risk accuracy,
under-triage, and clarification recall and specificity.
Category accuracy measures label agreement, not
factual correctness. Since 72 messages have labels
outside the supported schema, we report full-sample
($n=540$) and compatible-label ($n=468$) accuracy.
Under-triage denotes same-day cases predicted as
routine ($n=174$), excluding urgent-category errors.

All configurations produce 540 parse-valid outputs
under a common parser, with minor repairs for three
FT+RAG and two shuffled outputs. Paired 95\%
confidence intervals use 10,000 source-case bootstrap
replicates (seed 20261008), preserving all six
variations of each case \cite{field2007}.

\subsection{Consumer Study}

Thirty consenting adults evaluated the web
prototype: 25 in English and five in Arabic.
Participants were informed of its nonclinical
purpose and instructed to avoid identifiable
health information. No names or contact details
were collected. Twenty-five participants were
aged 18--34, four were aged 35--54, and one was 55 or older.
Education included bachelor's (15), master's (9),
college (5), and doctoral (1) qualifications.
Nineteen reported high chatbot familiarity,
16 used chatbots frequently, and 25 had previously
used them for health information. Gender was
not recorded.

Participants interacted with three configurations
(Table~\ref{tab:systems}): the base model (A),
domain-adapted model (B), and domain-adapted
model with retrieval (C).

\begin{table}[t]
\caption{Configurations in the consumer study.}
\label{tab:systems}
\centering
\footnotesize
\begin{tabular}{ll}
\toprule
\textbf{System} & \textbf{Configuration}\\
\midrule
A & Base Qwen3.5-9B with instructions\\
B & QLoRA-adapted Qwen3.5-9B\\
C & QLoRA-adapted Qwen3.5-9B + RAG\\
\bottomrule
\end{tabular}
\end{table}
\footnotetext[1]{The interactive prototype of the proposed system is available at: \url{https://herhealthgpt.org/}.}
\subsection{Questionnaire and Analysis}

The questionnaire assessed clarity, relevance,
usefulness, hesitation, follow-up quality,
perceived understanding, and confidence in sources.
It also recorded system preference, intended actions,
and qualitative feedback.

Most items used five-point scales; perceived
understanding retained its original six-point scale.
All 30 participants completed the ratings.
We report medians, quartiles, and counts of
ratings $\geq4$. System preference is summarized
using proportions and Wilson 95\% confidence
intervals.

Ratings describe the overall prototype rather
than individual systems. Only the preference
question compares A, B, and C. Since exposure
order was not recorded, no between-system
significance tests were performed. Arabic
responses are treated as exploratory.

\section{Results and Discussion}
This section presents the results of three complementary
evaluations: concern classification and risk assessment,
exploratory clarification routing, and consumer feedback.
We examine the contribution of evidence retrieval,
the trade-offs in conversational control, and user
perceptions of the implemented prototype.
\label{sec:results}

\subsection{Component Evaluation}

Table~\ref{tab:components} compares six model
configurations on 540 English messages.
FT+Real RAG achieves the highest concern-category
accuracy (59.3\%), compared with 54.1\% for FT.
Its paired improvement is 5.19 percentage points
(95\% CI [1.11, 9.26]). Improvements over
FT+Instruction and FT+Shuffled are 5.00
[1.11, 9.07] and 6.11 [2.22, 10.37] points,
respectively. All three post-hoc 98.33\%
intervals remain positive. Gains persist on
schema-compatible messages and after excluding
one case with identified training-text overlap.

\begin{table*}[t]
\caption{English component results. Category accuracy:
540 messages (full) and 468 (compatible); risk:
540; under-triage: 174; clarification recall:
90 and specificity: 450. All values are percentages.}
\label{tab:components}
\centering
\footnotesize
\setlength{\tabcolsep}{5pt}
\begin{tabular}{lrrrrrr}
\toprule
Configuration & Category & Compatible &
Risk & Under-triage $\downarrow$ &
Clar. recall & Clar. spec.\\
\midrule
ZS
&53.9&62.2&66.7&71.8&85.6&86.2\\
Base+RAG
&56.1&64.7&65.0&73.6&66.7&94.7\\
FT
&54.1&62.4&68.5&52.9&0.0&100.0\\
FT+Instruction
&54.3&62.6&67.0&58.6&0.0&100.0\\
FT+Shuffled
&53.1&61.3&67.6&59.2&0.0&100.0\\
\textbf{FT+Real RAG}
&\textbf{59.3}&\textbf{68.4}
&68.1&56.9&0.0&100.0\\
\bottomrule
\end{tabular}
\end{table*}

Classification gains do not establish better
urgency assessment. Relative to FT, FT+Real RAG
changes risk accuracy by $-0.37$ points
(95\% CI [$-4.26$, 3.33]) and under-triage
by +4.02 points [$-4.86$, 13.10].
Only 18 messages from three cases have urgent
references, of which FT+Real RAG correctly
predicts one as urgent. The adaptation-by-RAG
interaction is inconclusive (+2.96 points,
95\% CI [-1.67, 7.59]).

\subsection{Exploratory Clarification Routing}

All adapted configurations have zero recall
on the 90 constructed clarification targets.
An exploratory offline router therefore uses
the ZS clarification decision, returning its
question when positive and otherwise retaining
the FT+Real RAG response. The router identifies 77/90 positives
(85.6\% recall) while interrupting
62/450 negatives (86.2\% specificity).
Precision is 55.4\%, F1 is 67.2\%,
and 401/540 messages (74.3\%) proceed
directly to an answer.

The router was selected after benchmark
inspection, and its metrics are inherited
from ZS. A live wrapper was smoke-tested
on six variations of one case. These
exploratory results illustrate the trade-off
between clarification and interaction
burden but do not validate the deployed
conversational control loop.
\subsection{Consumer Evaluation}
\begin{table}[t]
\caption{Consumer ratings ($n=30$).
Counts indicate ratings $\geq4$.}
\label{tab:userresults}
\centering
\scriptsize
\setlength{\tabcolsep}{3pt}
\begin{tabular}{lccc}
\toprule
Measure & Median [Q1,Q3] &
EN & AR\\
\midrule
Easy to understand &4 [4,5]&21&5\\
Response relevant &4 [4,5]&20&5\\
Response useful &4 [4,5]&21&5\\
Hesitate to rely &3 [2,4]&7&1\\
Follow-ups relevant &4 [4,5]&22&3\\
Understood question$^{*}$ &5 [4,5]&21&2\\
Sources raised confidence &4 [4,5]&21&4\\
\bottomrule
\end{tabular}
\par\smallskip
\raggedright
\scriptsize
EN: $n=25$; AR: $n=5$.
$^{*}$Six-point scale; other items use
five levels. Agreement with hesitation
indicates caution.
\end{table}

Table~\ref{tab:userresults} presents questionnaire
results from 30 participants. Most participants rated the prototype
favorably for clarity, relevance, usefulness,
follow-up quality, and confidence in supporting
sources. However, reliance remained mixed:
eight participants expressed hesitation, while
thirteen disagreed.

\begin{figure}[t]
    \centering
    \begin{subfigure}[t]{0.49\columnwidth}
        \centering
        \includegraphics[width=\linewidth]{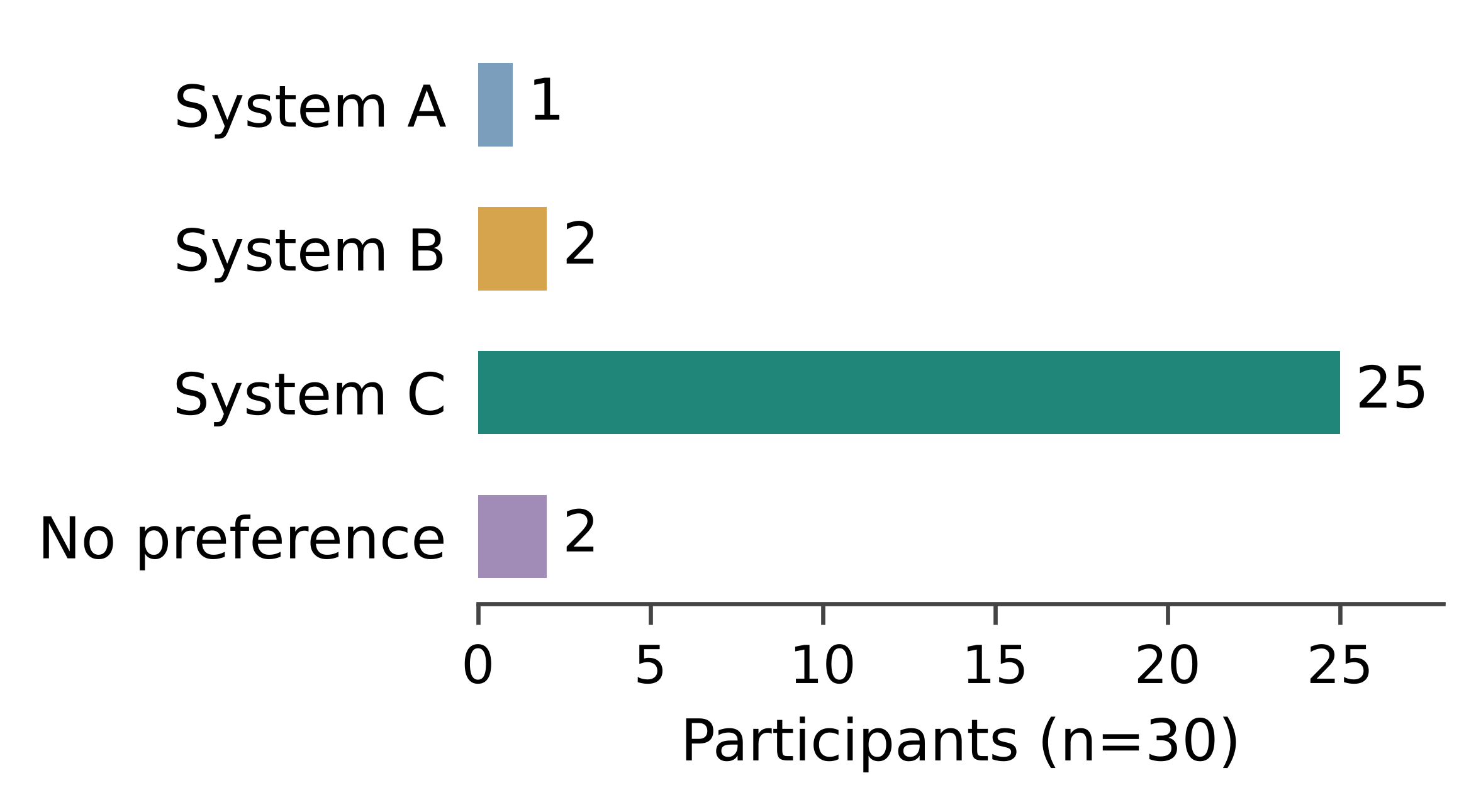}
        \caption{System preference.}
        \label{fig:consumer_preference}
    \end{subfigure}
    \hfill
    \begin{subfigure}[t]{0.49\columnwidth}
        \centering
        \includegraphics[width=\linewidth]{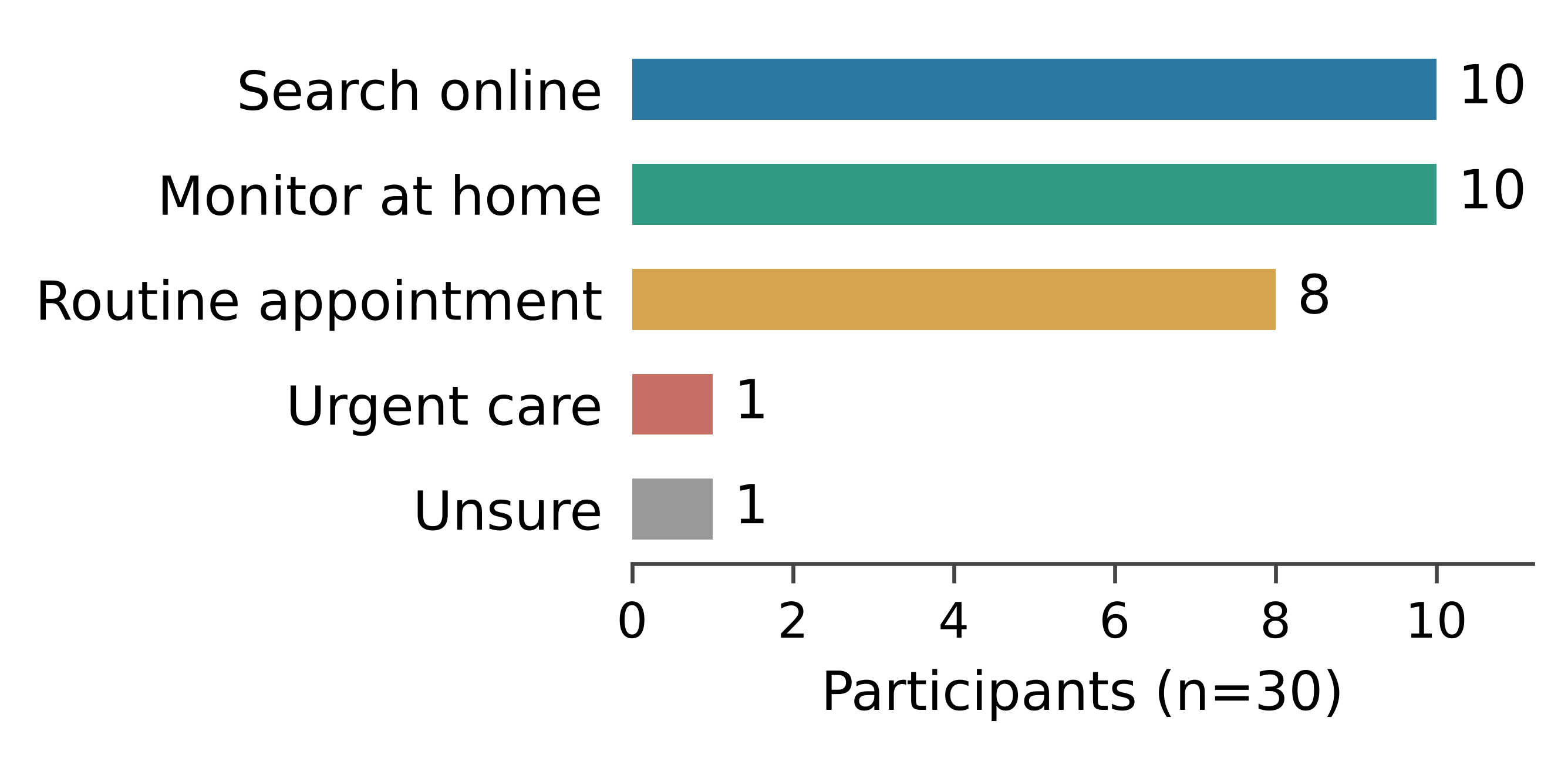}
        \caption{Intended next actions.}
        \label{fig:consumer_actions}
    \end{subfigure}

    \caption{Consumer's preferences.
    (n=30): (a) preferred system configuration
    and (b) intended actions.}
    \label{fig:consumer}
\end{figure}

Figure~\ref{fig:consumer} summarizes system
preferences and intended next actions.
System C, combining domain adaptation and
retrieval, was preferred by 25 participants
(83.3\%; Wilson 95\% CI [66.4, 92.7]),
followed by System B (2), System A (1),
and no preference (2). Participants favoring
System C highlighted follow-up questions,
perceived reliability, response speed,
and supporting sources.

As shown in Fig.~\ref{fig:consumer_actions},
participants most frequently selected online
searching (10), home monitoring (10), or routine
consultation (8), while one selected urgent
care and one was unsure.

Although preference favored System C,
the questionnaire did not collect separate
ratings for each configuration or record
exposure order. Therefore, the results
indicate favorable consumer perceptions
rather than an isolated retrieval effect.
Intended actions likewise represent
self-reported choices, not verified clinical
decisions.

\subsection{Discussion and Limitations}

Query-associated retrieval improves concern classification,
but does not establish better urgency assessment or clinical
reliability. Clarification routing shows promise but may
introduce unnecessary questions. The component evaluation
is limited by 90 source cases, heuristic labels, schema
mismatches, and potential training-data overlap. The small
questionnaire sample ($n=30$) and system-independent ratings
limit comparative conclusions, while five Arabic responses
cannot establish multilingual performance. Further work
should evaluate end-to-end conversational reliability,
clinical appropriateness, latency, and deployment performance.

\section{Conclusion}
\label{sec:conclusion}

This paper presented HerHealthGPT, an evidence-grounded
conversational AI framework for women's reproductive health
that integrates domain-adapted language models, retrieval,
and structured conversational control. Experimental results
demonstrate that relevant evidence improves concern
classification, achieving 59.3\% accuracy and a statistically
supported 5.19 percentage-point gain over fine-tuning alone.
A consumer study further revealed strong preference for
the retrieval-enabled configuration, selected by 83.3\%
of participants. However, the findings also highlight
critical challenges in urgency assessment and clarification,
underscoring that improved classification does not
guarantee clinically appropriate guidance. HerHealthGPT
therefore provides a practical foundation for developing
more transparent and context-aware reproductive-health
assistants while identifying the safeguards needed for
responsible deployment. Future work will prioritize
clinician-validated evaluation, safer conversational
decision-making, and robust multilingual support.

\end{document}